# AUTOMATIC CONSTRUCTION OF CLEAN BROAD-COVERAGE TRANSLATION LEXICONS

I. Dan Melamed [1]

Dept. of Computer and Information Science
University of Pennsylvania
Philadelphia, PA, 19104, U.S.A.
melamed@unagi.cis.upenn.edu

**Abstract**

Word-level translational equivalences can be extracted from parallel texts by surprisingly simple statistical techniques. However, these techniques are easily fooled by *indirect associations* — pairs of unrelated words whose statistical properties resemble those of mutual translations. Indirect associations pollute the resulting translation lexicons, drastically reducing their precision. This paper presents an iterative lexicon cleaning method. On each iteration, most of the remaining incorrect lexicon entries are filtered out, without significant degradation in recall. This lexicon cleaning technique can produce translation lexicons with recall and precision both exceeding 90%, as well as dictionary-sized translation lexicons that are over 99% correct.

## 1 Introduction

Translation lexicons are explicit representations of translational equivalence at the word level. They are central to any machine translation system, and play a vital role in other multilingual applications, including machine-assisted translation tools (e.g. [Mac95, Mel96b]), various methods of aligning parallel corpora (e.g. [DCG93, F&C94, Mel96a]), concordancing for bilingual lexicography [CRW89, G&C91], computer-assisted language learning, and multilingual information retrieval [MIR96]. Even some monolingual NLP tasks, such as word-sense disambiguation, can benefit from access to a translation lexicon [GCY93, D&I94].

An automatic method of constructing translation lexicons would be of great value, even if the method succeeded in automating only a part of the manual process. Yet, to my knowledge, nobody has devised an automatic method for constructing translation lexicons with high precision *and* broad coverage. I endeavor to remedy this situation in this paper.

There are many ways to organize a translation lexicon, but perhaps the most general representation is a set of ordered word pairs. This representation will be assumed throughout this paper. Formally, a **translation lexicon** for languages L1 and L2 is a subset of the crossproduct of the words of L1 and the words of L2. Each entry $E$ in the translation lexicon is an ordered pair $(v, w)$, where $v \in$ L1 and $w \in$ L2. Some automatic methods for producing translation lexicons attach probabilities, likelihoods, or confidence measures to each entry. I will refer to all such attributes as **association scores**. A high association score indicates that two words are strongly associated, by virtue of being "good" translations of each other in some sense of "good." A translation lexicon with an association score attached to each entry is called a **graded translation lexicon**.

---
[1] Thanks to Mickey Chandrasekar, Mike Collins, Jason Eisner, Mike Niv, Adwait Ratnaparkhi, Lyle Ungar and four anonymous reviewers for many helpful comments.

## 2 Previous Work

Several researchers have attempted automatic construction of translation lexicons, either as an end in itself, or as a stepping stone to other goals. Most prominent in the latter category is work on statistical machine translation [BD+90] and work on alignment of bilingual corpora [BLM91, DCG93]. Both research directions have been based on the IBM statistical translation models, which estimate $\Pr(t|s)$ for target words $t$ and source words $s$. These models are unsuitable for construction of translation lexicons for two reasons. First, for most source words $s$, the models maintain a long probabilistic "tail" of target words $t$, such that $\Pr(t|s)$ is very small but not zero. After being trained on a very large set of French/English parallel texts, one such model contained an average of 39 French translations for each English word [BD+90]. Second, if a source word $s$ has several correct translations $t$, then $\Pr(t|s)$ will be proportionally smaller for each $t$, since the sum of these probabilities must equal 1. The following pair of examples, taken from [BLM91] and [DCG93], demonstrate that these problems cannot be solved by clever thresholding.

| $e$ = "external" | | $f$ = "fermer" | |
|---|---|---|---|
| $f$ | $\Pr(f|e)$ | $e$ | $\Pr(e|f)$ |
| extérieures | 0.944 | close | 0.44 |
| extérieur | 0.015 | when | 0.08 |
| externe | 0.011 | Close | 0.07 |
| extérieurs | 0.010 | selected | 0.06 |

Most statistical algorithms aimed at producing translation lexicons between languages L1 and L2 (e.g. [CRW89, G&C91, Fun95, K&H94, Mel95, W&X95]) are variations on the following greedy algorithm:

1. Choose a similarity metric S between words in L1 and words in L2. The similarity metric is usually based on how often words co-occur in corresponding regions of a parallel text corpus, although other metrics have also been proposed [Fun95].

2. Compute association scores $S(v, w)$ for a set of word pairs $(v, w) \in$ (L1 × L2).

3. Sort the word pairs in descending order of their association score.

4. Choose a threshold $t$. The word pairs whose association score exceeds $t$ become the entries in the translation lexicon.

The greedy algorithm works remarkably well, considering how simple it is. The problem is that the association scores in Step 2 are typically computed independently of each other. The well-known problem with this independence assumption is illustrated in Figure 1. The sequences of $v$'s and $w$'s represent two aligned text segments. If $w_k$ and $v_k$ occur in similar bilingual contexts, then any reasonable similarity metric will produce a high association score between them, reflecting the interdependence of their distributions. If $w_k$ and $v_k$ are indeed mutual translations, then the association between them is called a **direct association**. Now, suppose that $v_k$ and $v_{k+1}$ appear in similar contexts. Then there is also a strong interdependence between the distributions of $v_k$ and $v_{k+1}$. Here is the problem: If $w_k$ and $v_k$ appear in similar contexts, and $v_k$ and $v_{k+1}$ appear in similar contexts, then $w_k$ and $v_{k+1}$ will also appear in similar contexts. The arrow connecting $w_k$ and $v_{k+1}$ in Figure 1 represents an **indirect association**, since the association between $w_k$ and $v_{k+1}$ arises only by virtue of the association between each of them and $v_k$. Algorithms that compute association scores independently of each other cannot tell the difference between a direct association and an indirect one. Not surprisingly, these algorithms produce translation lexicons full of indirect (and incorrect) associations.

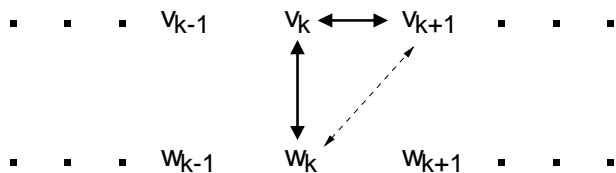

Figure 1: *The $v$'s and $w$'s are words in an aligned segment pair. The direct association between $w_k$ and $v_k$, and the association between $v_k$ and $v_{k+1}$ give rise to an indirect association between $w_k$ and $v_{k+1}$.*

Paradoxically, the irregularities (noise) in text and in translation mitigate the problem. If noise in the data reduces the strength of a direct association, then the same noise will reduce the strengths of any indirect associations that are based on this direct association. On the other hand, noise can reduce the strength of an indirect association without affecting any direct associations. Therefore, on average, direct associations are stronger than indirect associations. If all the entries in a translation lexicon are sorted by their associations scores, the direct associations will be very dense near the top of the list, and sparser towards the bottom.

Gale & Church [G&C91] have shown that entries at the very top of the list can be over 98% correct. They report lexicon entries for about 61% of the word tokens in a sample of 800 English sentences. To obtain 98% precision, their algorithm selected only entries for which it had high confidence that the association score was high. These would be the entries that co-occur most frequently. A random sample of 800 sentences from the same Hansards corpus showed that 61% of the word tokens, where the tokens are of the most frequent types, represent 4.5% of all the word types. A similar strategy was employed by Wu & Xia [W&X95] and by Fung [Fun95]. Wu & Xia constructed lexicon entries for 6517 English words from a 3.3-million-word corpus, with a precision of 86%. Fung skimmed off the top 23.8% of the noun-noun entries in her lexicon to achieve a precision of 71.6% for nouns only.[2]

It appears that the greedy thresholding algorithm has reached its limits. Incorrect lexicon entries must be removed in a more sophisticated way, in order to retain more of the correct entries in the final product.

## 3  How to Clean a Lexicon

The lexicon cleaning technique proposed here assumes an initial graded translation lexicon, and a corpus of aligned text segments. The aligned segments need not be aligned sentences; they can be the segments of any reasonably dense bitext map [Mel96a]. The initial translation lexicon can be constructed using any existing technique, but techniques that assign symmetric association scores are preferred. For example, likelihood ratios are symmetric, because the likelihood of words $v$ and $w$ being mutual translations is independent of which word came from the source language and which from the target language. In contrast, $\Pr(w|v)$ depends on how many valid translations exist for $v$, but not on how many translations exist for $w$. The lexicon cleaning technique presented here is completely symmetric[3]. Thus, words with multiple senses or with multiple morphological variants can participate in proportionally more lexicon entries.

---

[2]These two results should not be judged on the same scale, because it is arguably more difficult to construct translation lexicons between English and Chinese than between English and French.

[3]The words "source" and "target" are used only as labels.

The cleaning process is based on two simplifying assumptions. The **one-to-one assumption** is that each *instance* of each word translates to at most one other word *token*. So, the model does not allow one word to translate to a phrase or vice versa. The **no-synonyms assumption** is that synonymous word *types* never appear in the same text segment. Jointly, these two assumptions have the following implication: For each word token $v$, there will be at most one word $w$ in the parallel text segment for which $(v, w)$ is a correct lexicon entry.

If a segment containing $v$ is aligned with a segment containing $w$ and $w'$, then $(v, w)$ and $(v, w')$ should not both appear in the translation lexicon. If both do appear, then at least one is incorrect. Think of a competition among lexicon entries where there can be only one winner. If entry $(v, w)$ is a winner in some segment pair, then we say that $v$ and $w$ are **linked** in that segment pair. A reliable procedure for selecting the correct lexicon entry among several competitors can be a powerful lexicon cleaning agent. The tendency for direct associations to be stronger than indirect associations suggests the following heuristic: *The entry with the highest association score wins.*

In each pair of aligned segments, there can be as many competitions as there are words, and each candidate word pair can participate in two of them (one for each member of the pair). The outcome of the competitions will depend on the order of the competitions. So, we need a well-founded method to order the competitions within segment pairs. If we assume that stronger associations are also more reliable, then the order of the competitions is determined by the reliability of their winners. The procedure in each segment pair (S,T) is then quite simple:

1. Pick $v \in $ S and $w \in $ T, such that the lexicon entry $(v, w)$ has the highest possible association score. This entry would be the winner in any competition involving $v$ or $w$. So, consider $v$ and $w$ linked.

2. The one-to-one assumption implies that entries containing $v$ or $w$ cannot win any other competition in the segment pair (S,T). Therefore, remove $v$ from S and remove $w$ from T.

3. If there is another $v \in $ S and another $w \in $ T, such that $(v, w)$ is in the lexicon, then go to Step 1.

Now that we know how to link tokens in aligned segments, we can discuss how to convert token links into lexicon entries. We start with the following intuition: A lexicon entry that wins ten competitions and loses one is much more likely to be correct than an entry that loses ten and wins only one. If we assume that competitions in different segment pairs are independent, then the number of links for each lexicon entry will have a binomial distribution. Moreover, the correct entries and the incorrect entries will tend towards two different binomial distributions. Let $\lambda_{right}$ be the probability that any two words that are mutual translations are linked when they co-occur. Let $\lambda_{wrong}$ be the probability that any two words that are *not* mutual translations are linked when they co-occur. Note that $\lambda_{right}$ and $\lambda_{wrong}$ need not sum to 1 because they are conditioned on different events. Ideally, $\lambda_{right}$ should be 1 and $\lambda_{wrong}$ should be 0. Section 4 shows how to estimate the actual values from the distribution of links in the corpus.

Given $\lambda_{right}$ and $\lambda_{wrong}$, we can re-estimate the likelihood of a lexicon entry $E = (v, w)$ being correct. Let $n_E$ be the number of times that $v$ and $w$ co-occur in the corpus and let $k_E$ be the number of times that $v$ and $w$ are linked out of these $n_E$ co-occurrences. Let $B(k, n, p)$ be the probability of $k$ links, where $k$ has a binomial distribution with parameters $n$ and $p$. Then the likelihood ratio in favor of $E$ being correct is

$$L(E) = \frac{B(k_E, n_E, \lambda_{right})}{B(k_E, n_E, \lambda_{wrong})}. \tag{1}$$

Equation 1 can be used to regrade all the entries in the translation lexicon.

We now have all the tools we need to build an iterative translation lexicon cleaning algorithm:

1. Construct an initial graded translation lexicon.

2. Armed with the initial lexicon, return to the aligned corpus, and generate links in each pair of aligned segments.

3. Discard lexicon entries that are never linked.

4. Estimate $\lambda_{right}$ and $\lambda_{wrong}$ as shown in Section 4.

5. Regrade each lexicon entry, using Equation (1).

6. Go to Step 2 unless the lexicon reaches a fixed point or some other stopping condition is met.

## 4 Parameter Estimation

Each iteration of the cleaning algorithm above requires re-estimation of the model parameters $\lambda_{right}$ and $\lambda_{wrong}$. The most likely values of these parameters are those that maximize the probability of the model given the distribution of links in the corpus. To maximize this probability, we first appeal to Bayes' rule, and assume a uniform prior, so that $\Pr(model|data) \propto \Pr(data|model)$. Now, let $n_E$ be the number of times that the pair $E$ occurs in some aligned segment pair and let $k_E$ be the number of times that $E$ is linked out of these $n_E$ occurrences. If we assume that links for different entries are independent, then

$$\Pr(data|model) = \prod_E \Pr(k_E | n_E, \lambda_{right}, \lambda_{wrong}).$$

Let $\tau$ be the probability that two arbitrary co-occuring word tokens $v$ and $w$ are mutual translations. Then the probability that the pair $E$ will be linked $k_E$ times out of $n_E$ is a mixture of two binomials:

$$\Pr(k_E | n_E, \lambda_{right}, \lambda_{wrong}) = \tau B(k, n_E, \lambda_{right}) + (1 - \tau) B(k, n_E, \lambda_{wrong}).$$

The space of possible models can be simplified by expressing $\tau$ in terms of $\lambda_{right}$ and $\lambda_{wrong}$. Let $K$ be the total number of links in the corpus and let $N$ be the total number of co-occuring word token pairs: $K = \sum_E k_E$, $N = \sum_E n_E$. Let $\lambda$ be the probability that an arbitrary co-occuring pair of words will be linked, regardless of whether they are mutual translations. Since we can compute $K$ and $N$ directly, we can also compute $\lambda$ directly:

$$\lambda = K/N. \qquad (2)$$

At the same time, the two-binomial mixture model implies that

$$\lambda = \tau \lambda_{right} + (1 - \tau) \lambda_{wrong}. \qquad (3)$$

From Equations (2) and (3) and a little algebra we get:

$$\tau = \frac{K/N - \lambda_{wrong}}{\lambda_{right} - \lambda_{wrong}}.$$

Since $\tau$ is now a function of $\lambda_{right}$ and $\lambda_{wrong}$, only the latter two parameters need to be optimized. The optimization can be achieved in a number of ways. To save implementation time, I used the simplex method, starting from $\lambda_{right} = 1$ and $\lambda_{wrong} = 0$.

# 5 Experimental Method

To evaluate the lexicon cleaning technique, I used the same aligned Hansards corpus as Gale & Church [G&C91], except that I only used 300,000 aligned segment pairs to save time. The corpus was automatically pre-tokenized to delimit punctuation, possessive pronouns and elisions. Morphological variants in both halves of the corpus were stemmed to a canonical form.

The last pre-processing step was motivated by an observation made by Gale & Church [G&C91] in evaluating the precision of their lexicon construction method. Gale & Church note that "all but one of [the errors] involved a function word." More than 35% of the words in the English half of the Canadian Hansards are function words. Yet, the correct translation of a function word often depends more on its argument than on the function word itself. Catizone et al. [CRW89] provide a telling example for English translations of the German preposition *auf*: "*auf Erfolg hoffen* corresponds to *hope for success*, *auf dem Lande* to *in the country*, and *auf dem Tisch* to *on the table*." Because the distributions of function words are highly dependent on the function words' arguments, lexicon construction algorithms often mistakenly pair up function words with the translations of their arguments. Thus, function words account for a large portion of the errors in automatically constructed translation lexicons.

On the other hand, since the translations of function words are so unpredictable, function word entries in translation lexicons are of dubious utility. Even if some application truly requires such entries, it is not difficult to construct them by hand, since there will be at most a couple of hundred [CRW89]. Likewise, it is not difficult to construct a stop-list of function words. So, having constructed the requisite stop-lists, I followed the advice of Fung [Fun95] and deleted all function words from the corpus. Though I have yet to confirm the effect, I suspect that I traded a tiny loss in recall for a huge gain in precision.

An initial translation lexicon was constructed using the method in [Mel95] with no linguistic filters. The algorithm at the end of Section 3 was run until the model converged. Six iterations were required to reach this point. Table 1 shows some interesting changes at the end of each iteration. First, as expected, $\lambda_{right}$ increases while $\lambda_{wrong}$ decreases. Second, the first few iterations discard many entries. Third, the probability of the data given the model increases, because the data are easier to model when there is less noise. The mean entry log-likelihood is obtained by dividing the log-$Pr(data|model)$ by the number of entries. It shows that the increase in the $Pr(data|model)$ is not just a result of the lexicon shrinking.

Table 1: *Changes in the model parameters and in the lexicon after each cleaning iteration. The initial lexicon contained 330076 entries.*

| cleaning iteration | $\lambda_{right}$ | $\lambda_{wrong}$ | entries in lexicon | log of Pr( data \| model ) | mean entry log-likelihood |
|---|---|---|---|---|---|
| 1 | .969 | .00879 | 100601 | -942206 | -9.366 |
| 2 | .975 | .000402 | 88653 | -690935 | -7.794 |
| 3 | .975 | .0000501 | 88544 | -688065 | -7.771 |
| 4 | .975 | .0000236 | 88539 | -686950 | -7.759 |
| 5 | .975 | .00000471 | 88538 | -686750 | -7.757 |
| 6 | .975 | .00000442 | 88538 | -686651 | -7.755 |

The entries remaining after the last cleaning iteration were sorted by their likelihood scores. Figure 2 shows the distribution of the log-likelihood scores in the final lexicon on a log scale. The log scale helps to illustrate the plateaus in the curve. The longest plateau represents the set of word pairs that were linked once out of one co-occurrence (1/1) in the corpus. All these word pairs are equally likely to be correct. The second-longest plateau results from word pairs that were linked twice out of two co-occurrences (2/2) and the third longest plateau is from word pairs that were linked three times out of three co-occurrences (3/3). As usual, the entries with higher likelihood scores were more likely to be correct. By discarding entries with lower likelihood scores, recall can be traded off for precision. This trade-off was measured at three points, each representing a cutoff at the end of one of the three longest plateaus.

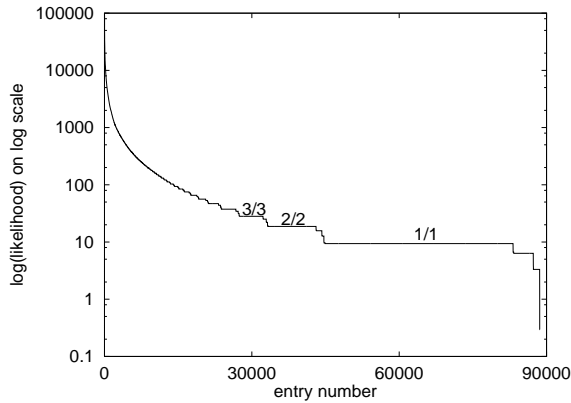

Figure 2: *The distribution of likelihood scores after the 2nd cleaning iteration. The long plateaus correspond to the most common combinations of $k_E/n_E$: 1/1, 2/2 and 3/3. All the entries E with the same $k_E/n_E$ combination are equally likely to be correct.*

## 6 Evaluation

Since translation lexicons consist of entries, the traditional method of measuring recall requires knowing what the right number of entries should be. However, this number depends on the *raison d'être* of the translation lexicon. A more widely applicable recall measure is based on the number of different words in the lexicon. For lexicons extracted from corpora, perfect recall implies at least one entry containing each word in the corpus. One-sided variants, which consider only source words, have also been used [G&C91]. Table 2 reports both the marginal (one-sided) and the combined recall at each of the three cut-off points. It also reports the number of entries in each of the three lexicons. Of course, the number of entries depends on the size of the bilingual corpus. Nevertheless, the numbers in Table 2 demonstrate the feasibility of automatically constructing translation lexicons as large as published bilingual dictionaries.

The next task was to measure precision. It would take too long to evaluate every lexicon entry manually. Instead, I took 5 random samples (with replacement) of 100 entries each from each of the three lexicons. Each of the samples was first compared to a translation lexicon extracted from a machine readable bilingual dictionary [CS+91]. All the entries in the sample that appeared in the dictionary were assumed to be correct. The remaining entries in all the samples were checked by hand.

Lexicon precision is a more thorny issue than lexicon recall. Human judges often disagree about the degree to which context should play a role in judgements of translational equivalence. The model of correctness used by judges who are not professional translators is the typical bilingual dictionary. Bilingual dictionaries are designed to answer the question, "What is the most likely set of translations for X?" For example, the Harper-Collins French Dictionary [CS+90] gives the following French translations for English "appoint": nommer, engager, fixer, désigner. Likewise, most lay judges would not consider "instituer" a correct French translation of "appoint." In actual translations, however, when the object of the verb is "commission," "task force," "panel," etc., the English "appoint" is often translated as "instituer" in French. In order to account for this kind of context-dependent translational equivalence, I evaluated the precision of the translation lexicons in the context of the parallel corpus whence they were extracted, using a simple bilingual concordancer. A lexicon entry $(v, w)$ was considered correct if $v$ and $w$ appeared as direct translations of each other in some aligned segment pair.

Table 2: *Lexicon recall at three different minimum likelihood thresholds. The bilingual corpus contained 41,028 different English words and 36,314 different French words, for a total of 77,342.*

| cut-off plateau | minimum likelihood score | total lexicon entries | English words represented | % | French words represented | % | total words represented | % |
|---|---|---|---|---|---|---|---|---|
| 3/3 | 28 | 32274 | 14299 | 35 | 13409 | 37 | 27708 | 36 |
| 2/2 | 18 | 43075 | 18533 | 45 | 17133 | 47 | 35666 | 46 |
| 1/1 | 9 | 88633 | 36371 | 89 | 33017 | 91 | 69388 | 90 |

Even direct translations come in different flavors. Most entries that I checked by hand were of the plain vanilla variety that you might find in a bilingual dictionary (entry type V). However, a significant number of words change their part of speech during translation (entry type P). For instance, in the entry (protection, protégé), the English word is a noun but the French word is an adjective. This entry appeared because "to have protection" is often translated as "être protégé" in the corpus. The entry will never occur in a bilingual dictionary, but users of translation lexicons, be they human or machine, will want to know that translations often happen this way.

Some lexicon entries contained direct translations that were incomplete (entry type I). For instance, one lexicon entry paired French "immédiatement" with English "right." As a one-to-one equivalence, this entry is incorrect. However, by analyzing the relevant aligned segment pairs in the corpus, I discovered that what the entry really captured was the association between "immédiatement" and "right away." The one-to-one pairing is the best that the algorithm could do, because it makes the simplifying assumption, as most others have done, that anything between consecutive spaces is a word. This assumption would be ludicrous for agglutinative languages, let alone Chinese or Japanese.

Still, there are three reasons to consider these incomplete entries correct. First, if the text corpus was preprocessed by an algorithm that correctly tokenized all the words in both languages, then the lexicon construction algorithm could never pair up incomplete words. If such preprocessing changes the overall precision, then it would be only for the better. Second, Smadja [Sma92] has shown how to expand incomplete words in translation lexicon entries, so that they become complete. Third, and most important, even incomplete entries are useful for many applications, such as concordancing for bilingual lexicography.

Table 3: *Distribution of different types of correct lexicon entries at varying levels of recall.*

| cutoff plateau | recall level | % type V | | % type P | | % type I | | total % precision | |
|---|---|---|---|---|---|---|---|---|---|
| | | mean | std. dev. | mean | std. dev. | mean | std. dev. | mean | std. dev. |
| 3/3 | 36% | 89 | 2.2 | 3.4 | 0.5 | 7.6 | 3.2 | 99.2 | 0.8 |
| 2/2 | 46% | 81 | 3.0 | 8.0 | 2.1 | 9.8 | 1.8 | 99.0 | 1.4 |
| 1/1 | 90% | 82 | 2.5 | 4.4 | 0.5 | 6.0 | 1.9 | 92.8 | 1.1 |

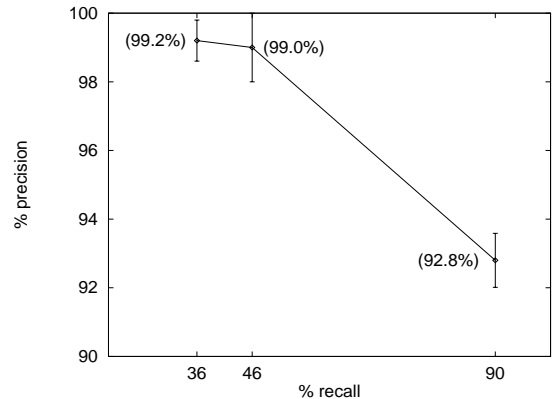

Figure 3: *Translation lexicon precision with 95% confidence intervals at varying levels of recall.*

Table 3 reports the distribution of correct lexicon entries among the types V, P and I. The three types were lumped together for the purposes of computing lexicon precision with 95% confidence intervals. These precision scores are plotted against recall in Figure 3.

## Conclusion

The main contribution of this work is a statistical method for cleaning noisy translation lexicons without significant degradation in recall. The method can generate translation lexicons with recall and precision both exceeding 90%, as well as dictionary-sized translation lexicons that are over 99% correct. Accurately measuring recall and precision of translation lexicons turned out to be more difficult than one might think.

The results presented here suggest several directions for future work. Previously, I have shown that the statistical construction of translation lexicons from parallel corpora can be gainfully assisted by incorporating various kinds of prior knowledge [Mel95]. I am optimistic that similar assistance can push the precision-recall envelope of the present method even higher. I am also eager to investigate methods for relaxing the one-to-one assumption, so that compound words can receive proper treatment. Most of all, I hope that the new availability of clean broad-coverage translation lexicons will fuel advances in NLP applications that rely on this resource.